\documentclass[10pt]{article}
\usepackage{graphicx}
\usepackage{epstopdf}
\usepackage{multirow}
\usepackage{cite}
\usepackage[cmex10]{amsmath}
\usepackage{amssymb}
\usepackage{amsmath,amssymb,amsthm,amsfonts,dsfont}
\usepackage{algorithmic}
\usepackage{algorithm}
\usepackage[tight,footnotesize]{subfigure}
\usepackage[font=footnotesize]{subfig}
\usepackage{url}
\usepackage{xspace}
\begin{document}

\title{An Efficient Genetic Algorithm for Discovering \textit{Diverse-Frequent} Patterns}

\author{Shanjida Khatun, Hasib Ul Alam and Swakkhar Shatabda\\Email: sshatabda@gmail.com}
\date{}
\maketitle

\begin{abstract}
Working with exhaustive search on large dataset is infeasible for several reasons. Recently, developed techniques that made pattern set mining feasible by a general solver with long execution time that supports heuristic search and are limited to small datasets only. In this paper, we investigate an approach which aims to find diverse set of patterns using genetic algorithm to mine diverse frequent patterns. We propose a fast heuristic search algorithm that outperforms state-of-the-art methods on a standard set of benchmarks and capable to produce satisfactory results within a short period of time. Our proposed algorithm uses a relative encoding scheme for the patterns and an effective twin removal technique to ensure diversity throughout the search.

\end{abstract}

\section{Introduction}

Recently pattern set mining has been used instead of pattern mining \cite{bringmann2010mining}. In pattern set mining, the aim is to find a small set of patterns in data that successfully partitions the dataset and discriminates the classes from one another \cite{guns2011declarative}. Many algorithms have been proposed in last few years to find such sets of patterns \cite{bringmann2010mining}. When the search space is too large or it is required to select a small set of patterns from a large dataset, exhaustive search techniques do not perform well.

Large data is challenging for most existing discovery algorithms because many variants of essentially the same pattern exist, due to (numeric) attributes of high cardinality, correlated attributes, and so on. While ignoring many potentially interesting results, this causes top-$k$ mining algorithms to return highly redundant result sets. These problems are particularly apparent with  pattern set discovery and its generalisation, exceptional model mining. To address this, we deal with the discriminative or diverse pattern set mining problem. We are given a set of transactions and a set of patterns in the concept learning set up to select a small set of  diverse patterns. In last few years, many algorithms that are proposed to solve the problem which are mostly exhaustive or greedy in nature \cite{guns2011declarative}. Constraint programming methods on a declarative framework \cite{guns2011declarative,guns2011itemset} have earned significant success. However, these algorithms perform very poorly for large datasets and requires huge time, where local search methods have been very effective to find satisfactory results efficiently.

We investigate the possibilities for studying diverse pattern sets to find small set of patterns within a short period of time using genetic algorithm with respect to a particular purpose by using a large datasets with minor modifications in the search technique. Our genetic algorithm has several novel components: a relative encoding technique learned from the structures in the dataset, a twin removal technique to remove identical and redundant individuals in the population and a random restart technique to avoid stagnation. We compared the performance with several other algorithms: random walk, hill climbing and large neighborhood search. The key contributions in the paper are as follows:

\begin{itemize}
\item Demonstrate the overall strength of our genetic algorithm for finding small set of diverse pattern.
\item Perform a comparative analysis between various types of local search algorithm and analysis of their relative strength compared with each other.
\end{itemize}

The paper is furnished as follows. In preliminaries section we explain our work and all the necessary definitions to understand the paper. In related work, we explain the previous task. In our approach part, we explain our algorithms and in experimental part, we explain our results and then conclude with a discussion and a possible outline for future work.
\section{Preliminaries}

\subsection{Pattern Constraints}

In this section, we explain some concepts to understand the diverse pattern set mining problems. These notations are adopted from Guns et al. \cite{guns2011declarative}.

 We assume that we are given a set of items $\mathcal{I}$ and a database, $\mathcal{D}$ of transactions $\mathcal{T}$, in which all elements are either 0 or 1. The process of finding the set of patterns which satisfy all of the constraints is called pattern set mining. A pair of variables $(I,T)$, where $I$ represents an itemset $I \subseteq \mathcal{I}$ and $T$ represents a transaction set $T \subseteq \mathcal{T}$ represented by means of boolean variables $I_i$ and $T_t$ for every item $i \in \mathcal{I}$ and every transaction $t  \in \mathcal{T}$. 

\begin{table}
\caption{A small example dataset containing five items and six transactions.\label{table1}}
\begin{center}
\begin{tabular}{clcccccc}
\hline
Transaction&&&&&\\
Id&ItemSet&A&B&C&D&E&Class\\
\hline
$t_1$&\{A,B,D\}&1&1&0&1&0&+\\
$t_2$&\{B,C\}&0&1&1&0&0&+\\
$t_3$&\{A,D\}&1&0&0&1&0&+\\
$t_4$&\{A,C,D\}&1&0&1&1&0&-\\
$t_5$&\{B,C,D\}&0&1&1&1&0&-\\
$t_6$&\{C,D,E\}&0&0&1&1&1&-\\
\hline
\end{tabular}

\end{center}
\end{table} 

The itemsets or pattern sets and the transaction sets are generally represented by binary vectors. The \textit{coverage} $\varphi_{\mathcal{D}}(I)$ of an itemset $I$ consists of all transactions in which the itemset occurs:
$$\varphi_{\mathcal{D}}(I)=\{t \in \mathcal{T}|\forall i \in I: \mathcal{D}_{ti}=1\}$$

For example, consider the small dataset presented in Table~\ref{table1}. Given an itemset, $I=\{B,C\}$, it is represented as $\langle 0,1,1,0,0 \rangle$ and the the coverage is $\varphi{\mathcal{D}}(I)=\{t_2,t_5\}$ which is represented by $\langle 0,1,0,0,1,0 \rangle$. Support of the itemset is $Support_{\mathcal{D}}(I)= 2$. Where, Support of an itemset is the size of its coverage set, $Support_{\mathcal{D}}(I)=|\varphi_{\mathcal{D}}(I)|$.

The dispersion score is the score of the frequent pattern sets based on the items categories within it. For example, for pattern set size, $k=3$, given three itemsets $I_1=\{B,C\}$, $I_2=\{C,D\}$ and $I_3=\{E\}$ in the pattern sets and their coverage will be $\varphi_{\mathcal{D}}(I_1)=\langle 0,1,0,0,1,0 \rangle$, $\varphi_{\mathcal{D}}(I_2)=\langle 0,0,0,1,1,1 \rangle$ and $\varphi_{\mathcal{D}}(I_3)=\langle 0,0,0,0,0,1 \rangle$ respectively. After XOR operation to each other, the sum of each item of the coverage will be

\begin{eqnarray*}
\varphi_{\mathcal{D}}(I_1) xor \varphi_{\mathcal{D}}(I_2)= \langle 0,1,0,1,0,1 \rangle =3,\\
\varphi_{\mathcal{D}}(I_1) xor \varphi_{\mathcal{D}}(I_3)= \langle 0,1,0,0,1,1 \rangle =3,\\
\varphi_{\mathcal{D}}(I_2) xor \varphi_{\mathcal{D}}(I_3) = \langle 0,0,0,1,1,0\rangle =2.\\
\end{eqnarray*} 

Now, the result of the dispersion score will be $3+3+2=8$.

\subsection{Pattern Set Constraints}

In pattern set mining, we are interested to find $k-$pattern sets \cite{guns2013k}. A $k-$pattern set $\Pi$ is a set of $k$ tuples, each of type $\langle I^p,T^p \rangle$. The pattern set is formally defined as following:
$$\Pi=\{\pi_1,\cdots,\pi_k\}, \text{ where, } \forall p =1,\cdots,k: \pi_p=\langle I^p,T^p \rangle$$ 
 
\textit{Diverse pattern sets}: In pattern set mining, highly similar transaction sets can be founded which can be undesirable. To avoid this, many measures can be used to find the similarity between two set of patterns such as dispersion score \cite{ruckert2007optimizing}:
\begin{eqnarray*}
dispersion(T^i, T^j)= \sum_{t\in\mathcal{T}}(2T_t^i -1)(2T_t^j -1).
\end{eqnarray*}
The term $(2T_t^i -1)$ transforms a binary $\{0,1\}$ variable into one of range $\{-1,1\}$. This way of finding dispersion score has some disadvantages. When two patterns cover exactly the same transactions and one pattern covers exactly the opposite transactions of the other, the score will be maximized in both. For example, if two patterns cover $\langle 0,1,1,0,0,1 \rangle$ and $\langle 1,0,0,1,1,0 \rangle$ or $\langle 0,1,1,0,0,1 \rangle$ and $\langle 0,1,1,0,0,1 \rangle$ transactions respectively, in both case, the score will be $6$ \cite{guns2011declarative}. This is not exactly desirable because in second case, it will must be $0$. To address this issue, we define and propose a new XOR based dispersion score to calculate the diversity between two pattern sets. 
\begin{eqnarray*}
xorDispersion(T^i, T^j)= \sum_{t\in\mathcal{T}}T_t^i \oplus T_t^j.
\end{eqnarray*}

To measure the diversity of a pattern set we use the following expression which is the objective function that we wish to maximize.
\begin{eqnarray*}
objDispersion=\sum_{i=1}^{k}\sum_{j=1}^{i-1}xorDispersion(T^i, T^j).
\end{eqnarray*}

To find diverse-frequent patterns, in last few years, most of the algorithms too struggles to produce good quality solutions on the large datasets within a short period of time. In this paper, to solve this problem, we proposed a XOR based genetic algorithm with various novel components which worked with large datasets.

\section{Related Work}
Many variants of pattern set mining are investigated in the literature. Among them to find patterns which are correlated \cite{rossi2006handbook}, discriminative \cite{shaw1998using}, contrast \cite{guns2013k} and diverse \cite{ruckert2007optimizing} became promising tasks.
Various algorithms has been proposed as a general framework for pattern mining \cite{guns2011declarative}, \cite{guns2011itemset} in last few years. Many languages have been developed for declaratively modeling problems, such as Zinc \cite{marriott2008design}, Essence \cite{frisch2008essence}, Gecode \cite{team2006gecode} and Comet \cite{guns2011declarative}, \cite{hentenryck2009constraint}. 

To search and prune the solution space, most of these methods use systematic search methods and the algorithms, those are not only exhaustive in nature but also take huge amount of time. On the other hand, stochastic search algorithms does not guarantee optimality but give a approximately best results within a short period of time. However, Guns
et al. \cite{guns2011declarative} investigated a technique by simplifying pattern set mining tasks and search strategies by putting these into a common declarative framework. In a recent work, Hossain et al. \cite{hossain2014stochastic} explored the use of genetic algorithms and other stochastic local search algorithms to solve the concept learning task using small datasets.

\section{Our Approach}
In this section, first we describe our proposed genetic algorithm to solve the diverse pattern set problem. Then we describe the other algorithms that we implemented in order to compare with our algortihm.

\subsection{Genetic algorithm} 

\begin{algorithm}
 \caption{geneticAlgorithm()}\label{Alg:ga}
\begin{algorithmic}[1]
\STATE $p=populationSize$
\STATE $percentChange=90$
\STATE $\mathcal{P}$ = generate $p$ valid pattern sets
\STATE $\mathcal{P}_{b}$ = \{\}
\WHILE {timeout}
\STATE $\mathcal{P}_m$ = simpleMutation($\mathcal{P}$)
\STATE $\mathcal{P}_c$ = uniformCrossOver($\mathcal{P}$)
\STATE $\mathcal{P}_*$ = select best ($\mathcal{P}\cup \mathcal{P}_m \cup \mathcal{P}_c$)
\IF{$\mathcal{P}_*$ remains same for 100 iteration}
\STATE $\prod$ = findBest($\mathcal{P}_{b}$)
\STATE $\mathcal{P}_{b}$ = $\mathcal{P}_{b}\cup\{\prod\}$ 
\STATE $\mathcal{P}_*$ = changePopulation( $percentChange,\mathcal{P}_*$)
\ENDIF
\STATE $\mathcal{P}=\mathcal{P}_*$ 
\ENDWHILE
\STATE $\prod^*$ = findBest($\mathcal{P}_{b}$)
\STATE \textbf{return} $\prod^*$
\end{algorithmic}
\end{algorithm}

Genetic algorithms are inspired by natural selection process. The search improves from generation to generation of a population of individuals by means of mutation and crossover. We have used XOR operation to generate our objective score as described in the preliminaries section.

In initialization part, we randomly generated  $p$ valid pattern sets and kept it in $\mathcal{P}$. To generate a valid pattern set we noticed that the itemsets have a particular structure. There are several exclusive attributes which are not true at a time. To avoid such invalid situations we used a constrained initialization for the representation.

Then we created population in $\mathcal{P}_{m}$ and $\mathcal{P}_{c}$. $\mathcal{P}_{m}$ created a population using mutation(shown in Algorithm~\ref{Alg:mutation}) and $\mathcal{P}_{c}$ created a population using cross over (shown in Algorithm~\ref{Alg:co}). After that we took best population from $\mathcal{P}$, $\mathcal{P}_{m}$ and $\mathcal{P}_{c}$ into $\mathcal{P}_*$. Here, size of $\mathcal{P}_*$ will be same as population size. We have iterated the procedure over and over again through several generations. If {$\mathcal{P}_*$ remains same for at least $100$ generations, we changed the value of $\mathcal{P}_*$ using $\textsf{simpleMutation}( PatternSets \mathcal{P})$ (shown in Algorithm~\ref{Alg:mutation}). This way we won't stuck in local minima. Here, We saved the maximum diverse pattern set in $\mathcal{P}_{b}$ every time. Then we copied $\mathcal{P}_*$'s value in $\mathcal{P}$. In the next generation, we got a new population. We continued this procedure until timeout. After that we returned the best score from $\mathcal{P}_{b}$.

We have checked the effect of population in result using \textit{tic-tac-toe} dataset. We have found that population size plays a pivotal role for generating result. We have described about this in analysis section elaborately.  
   
\begin{algorithm}
 \caption{simpleMutation( PatternSets $\mathcal{P}$)}\label{Alg:mutation}
\begin{algorithmic}[1]
\STATE $index=0$
\STATE $\mathcal{P}_{m}$ = \{\}
\STATE $size=$ noOfPatternset($\mathcal{P})$
\WHILE {$index$ $\textless$ $size$}
\STATE   $\prod$  $=$ $\mathcal{P}$[$index$]
\STATE  $\prod_m$  $=$ generate a valid neighbor of $\prod$ by flipping single bit
\WHILE {$\prod_m$ $\in$ $\mathcal{P}_{m}$}
\STATE $\prod_m$  $=$ generate a valid neighbor of $\prod$ by flipping single bit
\ENDWHILE
\STATE $\mathcal{P}_{m}$ = $\mathcal{P}_{m}\cup \{$ $\prod_m\}$  
\STATE $index++$
\ENDWHILE
\STATE \textbf{return} $\mathcal{P}_{m}$  
\end{algorithmic}
\end{algorithm}

Using $\textsf{simpleMutation}( PatternSets \mathcal{P})$, we have created $p$ new pattern sets by mutation. We have generated pattern sets randomly by changing a single bit. While doing the mutation, we always kept the structure constraint satisfied.     

\begin{algorithm}
 \caption{crossOver( PatternSets $\mathcal{P}$)}\label{Alg:co}
\begin{algorithmic}[1]
\STATE $index=0$
\STATE $\mathcal{P}_{c}$ = \{\}
\STATE $size=$ noOfPatternset($\mathcal{P})$
\WHILE {$index$ \textless $size$}
\STATE   $\prod_m$  $=$ randomly take a pattern set from $\mathcal{P}$
\STATE  $\prod_f$  $=$ randomly take a pattern set from $\mathcal{P}$
\STATE $\prod_o$  $=$ uniformCrossOver( $\prod_m$, $\prod_f$ )
\WHILE {$\prod_o$ $\in$ $\mathcal{P}_{c}$}
\STATE $\prod_o$  $=$ uniformCrossOver( $\prod_m$, $\prod_f$ )
\ENDWHILE
\STATE $\mathcal{P}_{c}$ = $\mathcal{P}_{c}\cup\{$$\prod_o\}$  
\STATE $index++$
\ENDWHILE
\STATE \textbf{return} $\mathcal{P}_{c}$  
\end{algorithmic}
\end{algorithm}

Using crossover (shown in Algorithm~\ref{Alg:co}), we have taken  two pattern sets from population to create an offspring. We have done this for $p$ times where $p$ is the number of population. Now we have $p$ offspring. We have used uniform crossover to find the offspring. We have randomly chosen each item from these two pattern set and place them into  new pattern sets but we have made sure that no duplicate remains in new population. The structure constraint is also satisfied during crossover.

\begin{algorithm}
 \caption{changePopulation($perChange$, PatternSets $\mathcal{P}$)}\label{Alg:changePopulation}
\begin{algorithmic}[1]
\STATE $noOfchange=(perChange$ $*$ sizeOf$(\mathcal{P}))/100$
\STATE remove lowest $noOfchange$ $\prod$ from $\mathcal{P}$
\STATE $i=1$
\WHILE {$i$ $\leq$ $noOfchange$}
\STATE $\prod$ = randomly create a valid patten set with $k$-items 
\WHILE {$\prod$ $\in$ $\mathcal{P}$}
\STATE $\prod$ = randomly create a valid patten set with $k$-items
\ENDWHILE
\STATE $\mathcal{P}$ = $\mathcal{P}\cup\{\prod\}$  
\STATE $i++$
\ENDWHILE
\STATE \textbf{return} $\mathcal{P}$ 
\end{algorithmic}
\end{algorithm}

To avoid getting stuck in local minima, we have used random restart in our genetic algorithm. When list of population aren't change for a certain period, we restarted the algorithm based on two variable. One, when it will be  restarted, and second, how much change will be done in the list. $\textsf{changePopulation}( percentChange, \mathcal{P} )$ (shown in Algorithm~\ref{Alg:changePopulation}) is used to create a new population where $\mathcal{P}$ represents the pattern set in which we have to change. $percentChange$ represents how much patters that we have to change. For example if $percentChange=90$, that means $90\%$ value will be deleted to create new value. In our algorithm, we experimented with different values of $percentChange$. We have found that when $percentChange=90$, we have always got good results. As it saves only top $10\%$ score and other $90\%$ will be used to create new population. 

In our Algorithm, we never allowed it to have twin in any population. Before entering any pattern sets, we have checked that if it is twin or not. When  it was already in there, we rejected it and created new one. We have done this until found a distinct valid pattern set. 

To find the objective score for a pattern set, we found coverage of each itemset. This will return some boolean array. After that we found all the combination for those boolean array. Now for each combination, we used XOR operator and added all the values.

\begin{algorithm}
 \caption{largeNeigbourhoodSearch()}\label{Alg:lns}
\begin{algorithmic}[1]
\STATE $noOfBitToChange=1$ 
\STATE $\prod$ = randomly create a valid patten set with $k$-items
\WHILE {timeout}
\STATE $\mathcal{P}$ = create $2^{noOfBitToChange}$ neighbours for $\prod$
\STATE  $\prod^\ast$ = find best individual from $\mathcal{P}$
\IF{getObjectiveScore( $\prod^\ast$ ) $\textgreater$  getObjectiveScore( $\prod$ )}
\STATE $ \prod = \prod^\ast$
\ENDIF
\IF{$\prod$  remains same for 100 iteration}
\STATE $noOfBitToChange++$
\ENDIF
\ENDWHILE
\STATE \textbf{return} $\prod$ 

\end{algorithmic}
\end{algorithm}

\subsection{Large Neighborhood Search(LNS)}
A large neighborhood search (LNS) is also implemented following the implementation in \cite{guns2011declarative}. For LNS (shown in Algorithm~\ref{Alg:lns}), we first created a valid pattern set and found its score. Then we created  its neighbors  and  found the best neighbor. If best neighbor is greater than the initial pattern set then we changed the initial pattern set and replaced it with best neighbor. In our implementation, the number of neighbors created for a pattern set will be $2^n$ where $n= noOfBitToChange$. When we generated the neighbors, at first we created $2^1$ neighbor with $n=1$. If it didn't give good results for $100$ iteration, we incremented the value of $n$ by $1$. We perform this again and again whenever LNS stuck for 100 iteration. To crate neighbors of a pattern set, we randomly choose an itemset from that pattern set. After that we randomly choose an item from that itemset. We do this for $n$ times as each item is represented by boolean values so if we creates all posssible neighbors for three items then number of neighbors for changing three items will be $2^3$. So, for $n$, it will be $2^n$.     
\begin{algorithm}
 \caption{hillClimbing()}\label{Alg:hc}
\begin{algorithmic}[1]
\STATE $\prod^\ast$ = randomly create a valid patten set with $k$-items
\STATE $bestScore$ = getObjectiveScore( $\prod^\ast$ ) 
\WHILE {timeout}
\STATE $\prod$ = generate a valid neighbor from $\prod^\ast$
\STATE  $currentScore$ = getObjectiveScore( $\prod$ ) 
\IF{$currentscore$ $\textgreater$ $bestScore$}
\STATE $\prod^\ast$ = $\prod$
\STATE $bestScore = currentScore$
\ENDIF
\ENDWHILE
\STATE \textbf{return} $\prod^\ast$ 
\end{algorithmic}
\end{algorithm}

\subsection{Hill Climbing with Single Neighbor}
For hill climbing (shown in Algorithm~\ref{Alg:hc}), we created a valid pattern set $\prod^\ast$ and copied the value of it in another pattern set called $\prod$. We started a loop which run for $1$ minute. Then we created a neighbor of $\prod^\ast$ in $\prod$. If this new neighbor is greater than the $\prod^\ast$, we copied the value of new neighbor in $\prod^\ast$ and created a new neighbor of $\prod^\ast$. The cycle goes on until the time is up.

\begin{algorithm}
 \caption{randomWalk()}\label{Alg:rw}
\begin{algorithmic}[1]
\STATE $bestScore = -\infty$
\STATE $\prod^\ast$ = $\phi$ 
\WHILE {timeout}
\STATE $\prod$ = randomly create a valid patten set with $k$-items
\STATE  $currentScore$ = getObjectiveScore( $\prod$ ) 
\IF{$currentscore$ $\textgreater$ $bestScore$}
\STATE $\prod^\ast = \prod$
\STATE $bestScore = currentScore$
\ENDIF
\ENDWHILE
\STATE \textbf{return} $\prod^\ast$ 
\end{algorithmic}
\end{algorithm}

\subsection{Random Walk}

In random walk (shown in Algorithm~\ref{Alg:rw}), we created a valid pattern set $\prod$. Then we created another pattern set called $\prod^\ast$. We copied the value of $\prod$ into $\prod^\ast$. Then we started a loop which run for $1$ minute. Here, we changed the $\prod$ by creating a new valid pattern set and then checked the value with $\prod^\ast$. If the score of $\prod$ is greater, we copied $\prod$ into $\prod^\ast$. Then again we changed $\prod$ by creating another pattern set randomly. This procedure is worked for $1$ minute. After that we took the score of $\prod^\ast$.

\section{Experimental Results}

 We have implemented all algorithms in JAVA language and have run our experiments on an Intel core i3 2.27 GHz machine with 4 GB ram running 64bit Windows 7 Home Premium.

\begin{table}[h]
\caption{Description of datasets.}
\scriptsize
\centering
\begin{tabular}{|c|c|c|}
\hline
\textbf{Data Set} & \textbf{Items}  & \textbf{Transactions}  \\
\hline Tic-tac-toe & 27 &  958  \\
Primary-tumor& 31 & 336  \\
Soybean & 50 & 630  \\
Hypothyroid & 88 & 3247 \\
Mushroom & 119 & 8124 \\
\hline
\end{tabular}
\label{table:1}
\end{table}

\subsection{Dataset}
In this paper,  the datasets that we use are taken from UCI Machine Learning repository \cite{frank2010uci} and originally used in \cite{guns2011declarative}. The datasets are available to download freely from the website: https://dtai.cs.kuleuven.be/CP4IM/datasets/. The datasets are given in Table \ref{table:1} with their properties.

\begin{table*}[t]
\centering
\caption{Objective score achieved by different algorithms for various datasets with different sizes of pattern sets $k$.}
\label{table:2}
\resizebox{\textwidth}{!}{%
\begin{tabular}{|c|c|c|c|c|c|c|c|c|c|}
\hline
\multirow{3}{*}{Data set} & \multirow{3}{*}{\begin{tabular}[c]{@{}c@{}}Pattern set size\\ k\end{tabular}} & \multicolumn{8}{c|}{Search Algorithm} \\ \cline{3-10} 
 &  & \multicolumn{2}{c|}{Random walk} & \multicolumn{2}{c|}{Hill Climbing} & \multicolumn{2}{c|}{LNS} & \multicolumn{2}{c|}{Genetic Algorithm} \\ \cline{3-10} 
 &  & Avg. & Best & Avg. & Best & Avg. & Best & Avg. & Best \\ \hline
Tic-tac-toe & \begin{tabular}[c]{@{}c@{}}2\\ 3\\ 6\\ 9\\ 10\end{tabular} & \begin{tabular}[c]{@{}c@{}}771\\ 1491.4\\ 5355\\ 17517.6\\ 11393.8\end{tabular} & \begin{tabular}[c]{@{}c@{}}\textit{\textbf{798}}\\ 1690\\ 5380\\ 18224\\ 12764\end{tabular} & \begin{tabular}[c]{@{}c@{}}516.8\\ 1432.2\\ 7004.4\\ 15977.6\\ 19963\end{tabular} & \begin{tabular}[c]{@{}c@{}}753\\ 1593\\ 7653\\ 16972\\ 21496\end{tabular} & \begin{tabular}[c]{@{}c@{}}762\\ 1825.6\\ 7758\\ 18097.6\\ 22235.2\end{tabular} & \begin{tabular}[c]{@{}c@{}}\textit{\textbf{798}}\\ \textit{\textbf{1916}}\\ 7791\\ 17858\\ 22748\end{tabular} & \begin{tabular}[c]{@{}c@{}}\textbf{798}\\ \textbf{1916}\\ \textbf{7938}\\ \textbf{18458.4}\\ \textbf{22731.4}\end{tabular} & \begin{tabular}[c]{@{}c@{}}\textit{\textbf{798}}\\ \textit{\textbf{1916}}\\ \textit{\textbf{7938}}\\ \textit{\textbf{18624}}\\ \textit{\textbf{22816}}\end{tabular} \\ \hline
Mushroom & \begin{tabular}[c]{@{}c@{}}2\\ 3\\ 6\\ 9\\ 10\end{tabular} & \begin{tabular}[c]{@{}c@{}}3388\\ 6889.6\\ 27260\\ 33955.2\\ 34117.2\end{tabular} & \begin{tabular}[c]{@{}c@{}}4936\\ 14576\\ 37440\\ 43216\\ 46584\end{tabular} & \begin{tabular}[c]{@{}c@{}}0\\ 3249.6\\ 0\\ 20960\\ 28868.4\end{tabular} & \begin{tabular}[c]{@{}c@{}}0\\ \textit{\textbf{16248}}\\ 0\\ 63392\\ 73116\end{tabular} & \begin{tabular}[c]{@{}c@{}}1362.4\\ 2070.4\\ 0\\ 0\\ 0\end{tabular} & \begin{tabular}[c]{@{}c@{}}6812\\ 10352\\ 0\\ 0\\ 0\end{tabular} & \begin{tabular}[c]{@{}c@{}}\textbf{8124}\\\textbf{ 16248}\\ \textbf{58734}\\ \textbf{103932}\\ \textbf{107529.6}\end{tabular} & \begin{tabular}[c]{@{}c@{}}\textit{\textbf{8124}}\\ \textit{\textbf{16248}}\\ \textit{\textbf{64992}}\\ \textit{\textbf{142452}}\\ \textit{\textbf{130944}}\end{tabular} \\ \hline
Hypothyroid & \begin{tabular}[c]{@{}c@{}}2\\ 3\\ 6\\ 9\\ 10\end{tabular} & \begin{tabular}[c]{@{}c@{}}439.6\\ 937.2\\ 2277\\ 3732.8\\ 5916.6\end{tabular} & \begin{tabular}[c]{@{}c@{}}562\\ 1484\\ 3405\\ 5864\\ 9333\end{tabular} & \begin{tabular}[c]{@{}c@{}}324.4\\ 0\\ 0\\ 0\\ 11689.2\end{tabular} & \begin{tabular}[c]{@{}c@{}}1622\\ 0\\ 0\\ 0\\ \textit{\textbf{29223}}\end{tabular} & \begin{tabular}[c]{@{}c@{}}649.4\\ 0\\ 0\\ 5193.6\\ 0\end{tabular} & \begin{tabular}[c]{@{}c@{}}\textit{\textbf{3247}}\\ 0\\ 0\\ 25968\\ 0\end{tabular} & \begin{tabular}[c]{@{}c@{}}\textbf{2736.4}\\ \textbf{5876}\\\textbf{ 12549.4}\\ \textbf{24234.8}\\ \textbf{17629.8}\end{tabular} & \begin{tabular}[c]{@{}c@{}}\textit{\textbf{3247}}\\ \textit{\textbf{6494}}\\ \textit{\textbf{16325}}\\\textit{\textbf{ 27556}}\\ 21726\end{tabular} \\ \hline
Soybean & \begin{tabular}[c]{@{}c@{}}2\\ 3\\ 6\\ 9\\ 10\end{tabular} & \begin{tabular}[c]{@{}c@{}}624\\ 1242.4\\ 3155\\ 5246.8\\ 6409\end{tabular} & \begin{tabular}[c]{@{}c@{}}624\\ 1248\\ 3438\\ 5778\\ 7597\end{tabular} & \begin{tabular}[c]{@{}c@{}}0\\ 260.4\\ 3304.2\\ 3770\\ 9406.2\end{tabular} & \begin{tabular}[c]{@{}c@{}}0\\ 1136\\ 5076\\ 5634\\ 12000\end{tabular} & \begin{tabular}[c]{@{}c@{}}374.5\\ 1168.8\\ 4023.8\\ 11113.6\\ 7653.8\end{tabular} & \begin{tabular}[c]{@{}c@{}}624\\ 1248\\ 4992\\ 12568\\ 12090\end{tabular} & \begin{tabular}[c]{@{}c@{}}\textbf{630}\\ \textbf{1260}\\ \textbf{5642.8}\\ \textbf{12547.2}\\ \textbf{15531.2}\end{tabular} & \begin{tabular}[c]{@{}c@{}}\textit{\textbf{630}}\\ \textit{\textbf{1260}}\\ \textit{\textbf{5664}}\\ \textit{\textbf{12598}}\\ \textit{\textbf{15696}}\end{tabular} \\ \hline
Primary-tumor & \begin{tabular}[c]{@{}c@{}}2\\ 3\\ 6\\ 9\\ 10\end{tabular} & \begin{tabular}[c]{@{}c@{}}326.4\\ 647.6\\ 2115.8\\ 3833.2\\ 4539\end{tabular} & \begin{tabular}[c]{@{}c@{}}329\\ 658\\ 2453\\ 4372\\ 4897\end{tabular} & \begin{tabular}[c]{@{}c@{}}238\\ 540.4\\ 2944\\ 6616.4\\ 7576.2\end{tabular} & \begin{tabular}[c]{@{}c@{}}\textit{\textbf{336}}\\ \textit{\textbf{672}}\\ 3017\\ 6710\\ 8336\end{tabular} & \begin{tabular}[c]{@{}c@{}}334.6\\ \textbf{672}\\ 3001.4\\ 6682\\ 8343.4\end{tabular} & \begin{tabular}[c]{@{}c@{}}\textit{\textbf{336}}\\ \textit{\textbf{672}}\\ 3018\\ 6712\\ \textit{\textbf{8393}}\end{tabular} & \begin{tabular}[c]{@{}c@{}}\textbf{336}\\ \textbf{672}\\ \textbf{3013.6}\\ \textbf{6715.2}\\ \textbf{8351.4}\end{tabular} & \begin{tabular}[c]{@{}c@{}}\textit{\textbf{336}}\\ \textit{\textbf{672}}\\ \textit{\textbf{3024}}\\ \textit{\textbf{6720}}\\ 8376\end{tabular} \\ \hline
\end{tabular}
}
\end{table*}

\subsection{Results}
In our experiment, we have implemented four algorithms. We have calculated the objective score for each algorithm. For each algorithm, we have used five datasets whose transaction number and item size can be found in Table \ref{table:1}. We have used $k$ pattern sets in each of them where $k = 2, 3, 6, 9, 10$. We have run each of them for $1$ minute and collected the score. For each test case, we have run the code five times and took its best score and average score. Which can be found in Table \ref{table:2}. We have found that almost all time genetic algorithm works better than other algorithms. In few cases, LNS works better as same as genetic algorithm. Random walk performs poorly. However, in few cases, hill climbing works better. 

\subsection{Analysis}
When number of itemset becomes greater, genetic algorithm prevails. In genetic algorithm, population size have to be in a limit. Too less or too many will give a bad result. Using random restart in genetic algorithm, changing $90\%$ population will work better.

Fig. \ref{fig:1} shows the effect of population size for the dataset \textit{tic-tac-toe}. We examined with different population size from $10-2000$. For each population, we ran the code five times and took best and average score. In X-axis, we put the population size and Y-axis we put the objective score. Fig. \ref{fig:1}(a) shows the average of objective score. In this figure, we can see that when population size is in $40-500$ it'll give the best answer. After that when population size is exceed $500$, the objective score will decrease. Fig. \ref{fig:1}(b) shows the best score. 
In this figure, we can see that when population size is in $10-1000$ it'll give the best answer. After that when population size is exceed $1000$, the objective score will decrease. So, we can conclude that genetic algorithm works more better with respect to population size but when the size of  population is small or big, we didn't get feasible answer in our allocated time since the calculations become too expensive. 

Fig. \ref{fig:2} shows the performance of the search algorithms base on their average objective score, which are shown as vertical bars, in $1$ minute  for all the datasets for different pattern set sizes. Here, genetic algorithm always gives good result with respect to other algorithms. Sometimes LNS gives good result as same as genetic algorithm. For the datasets \textit{mushroom} and \textit{hypothyroid}, the objective score of LNS and hill climbing becomes zero because the size of the items of the datasets (shown in Table \ref{table:1}) is too big. From Fig. \ref{fig:2} we also shows that hill climbing performs better than random walk which performance is very poor.

In Fig. \ref{fig:3}, we depict the performance of different search algorithms for the \textit{tic-tac-toe} dataset. 
In this figure, objective score of the search algorithms are shown as vertical line for different times. Random walk performs poorly as usual. However, hill climbing improves very quickly using single neighbor. LNS performs very well which result is near to genetic algorithm. However, genetic algorithm continuous gives best result. 

\begin{figure}[h]
\centering
\subfigure[Average]{\includegraphics[width=0.49\textwidth]{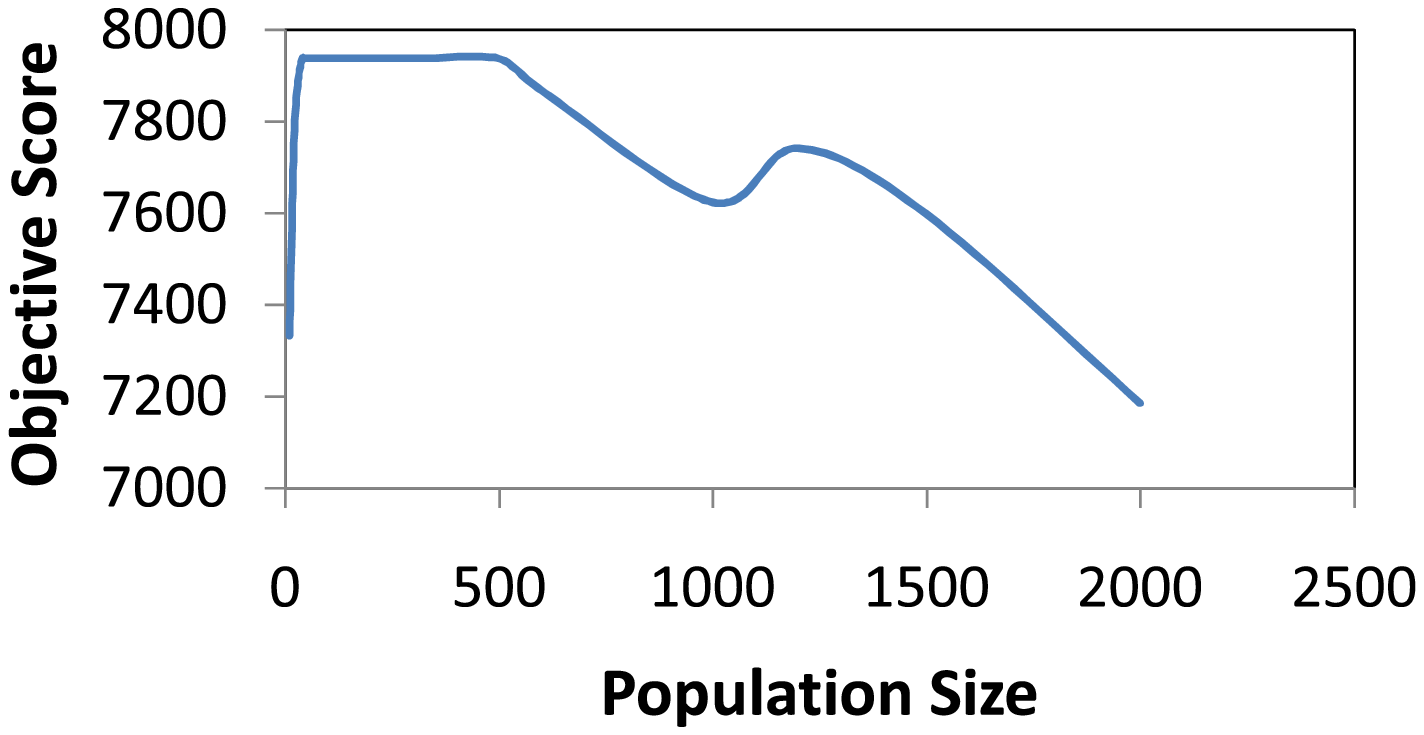}}
\subfigure[Best]{\includegraphics[width=0.49\textwidth]{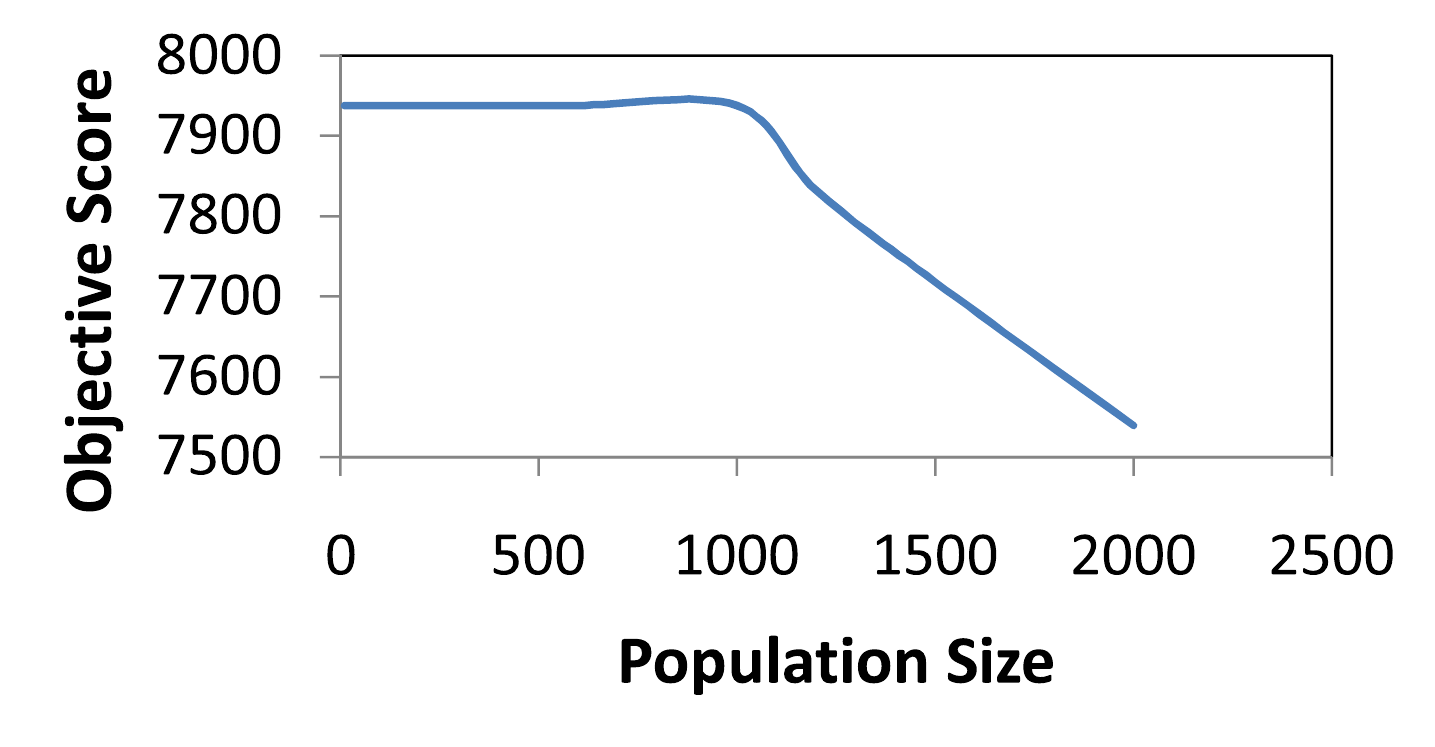}}
\caption{Search progress for genetic algorithm for the tic-tac-toe dataset with pattern size $k=6$. }
\label{fig:1}
\end{figure}

\begin{figure}[h]
\centering
\subfigure{\includegraphics[width=0.33\textwidth]{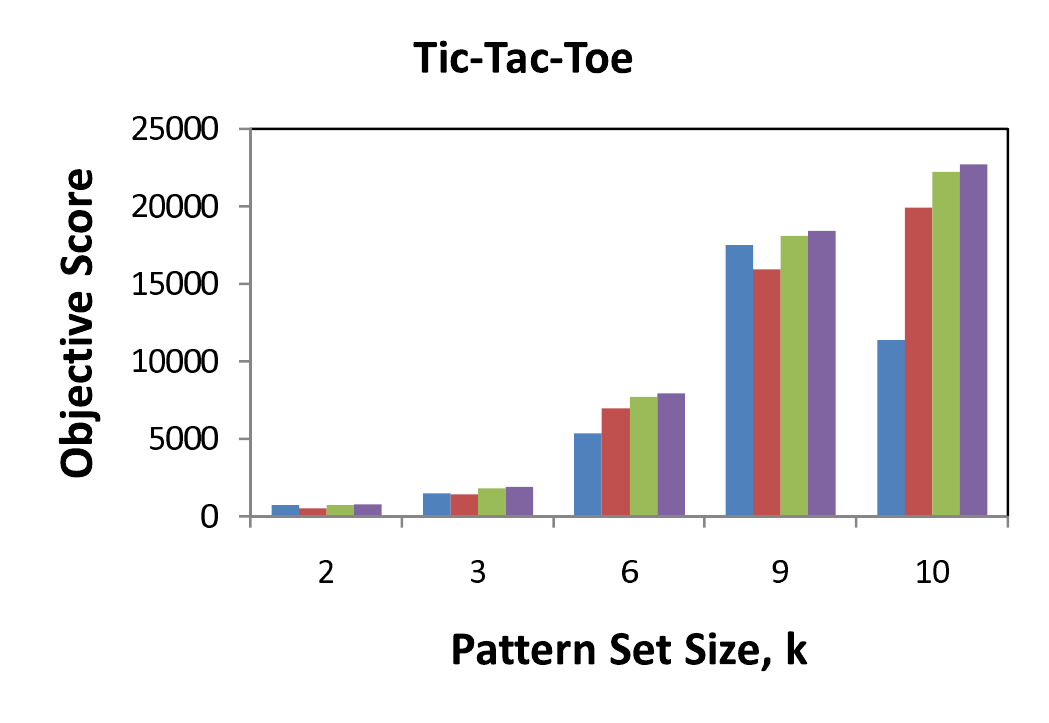}}
\subfigure{\includegraphics[width=0.33\textwidth]{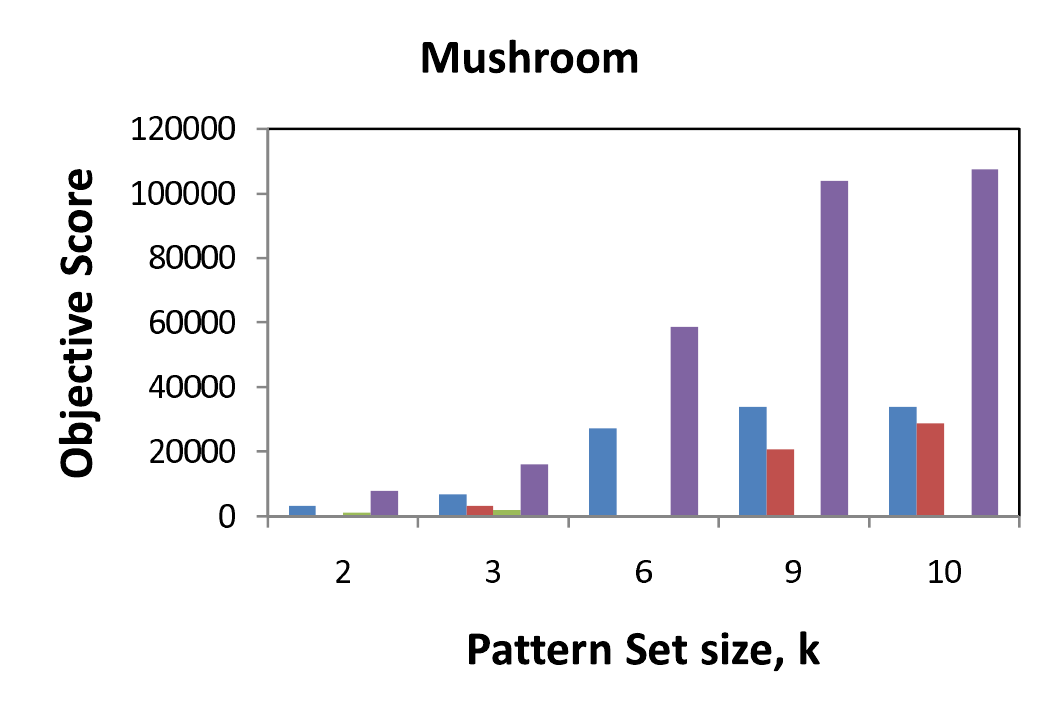}}
\subfigure{\includegraphics[width=0.33\textwidth]{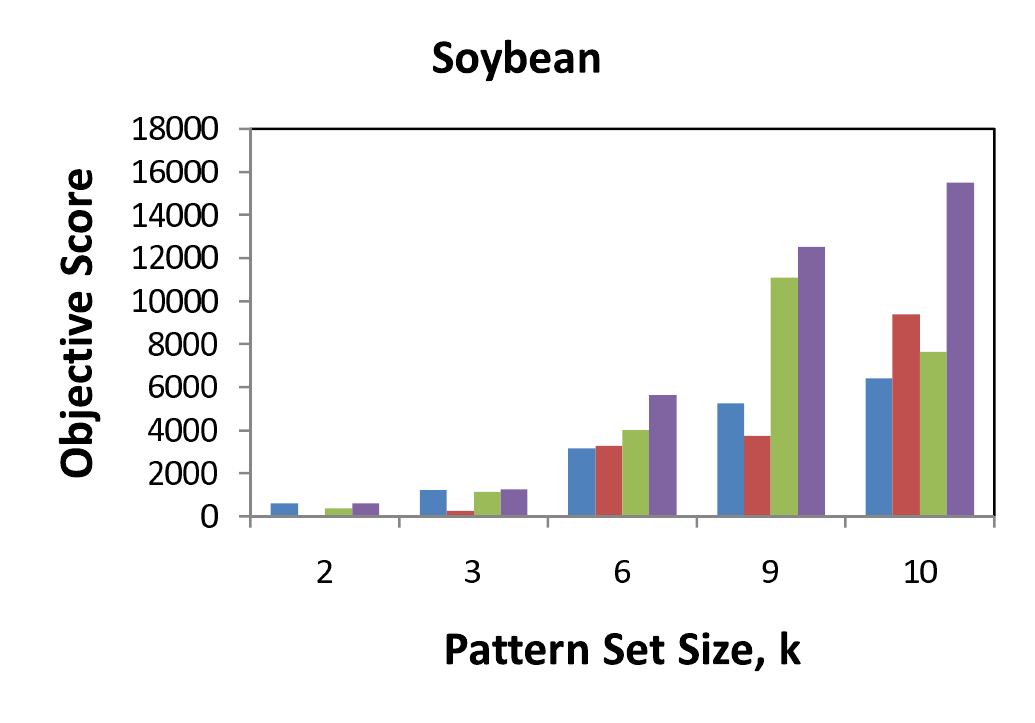}}
\subfigure{\includegraphics[width=0.33\textwidth]{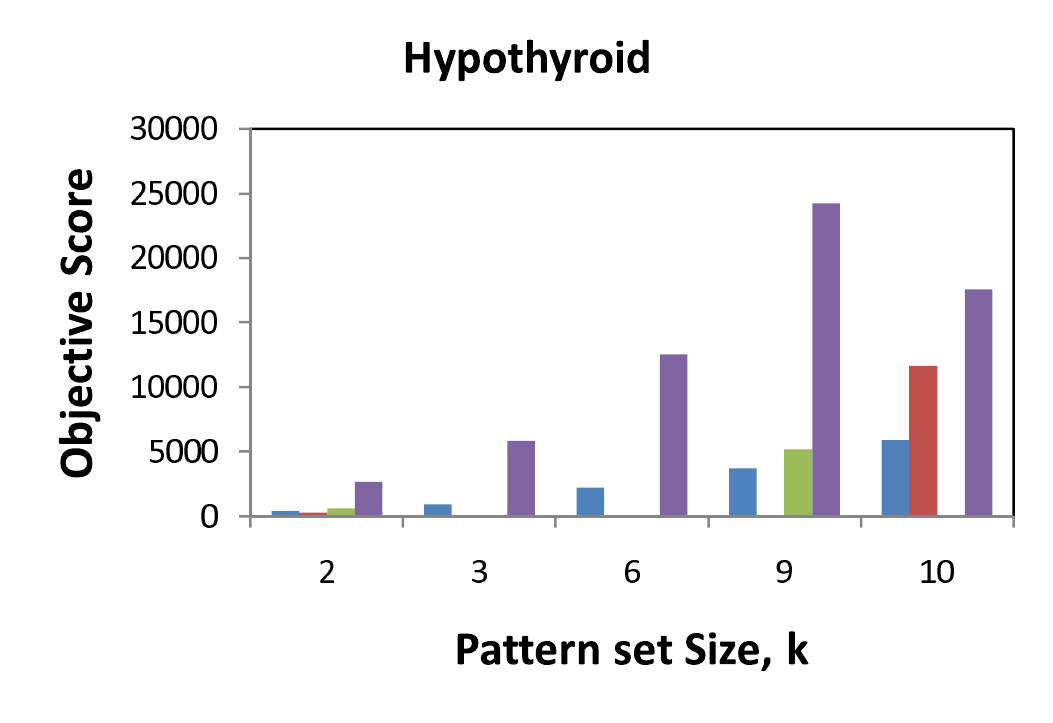}}
\subfigure{\includegraphics[width=0.33\textwidth]{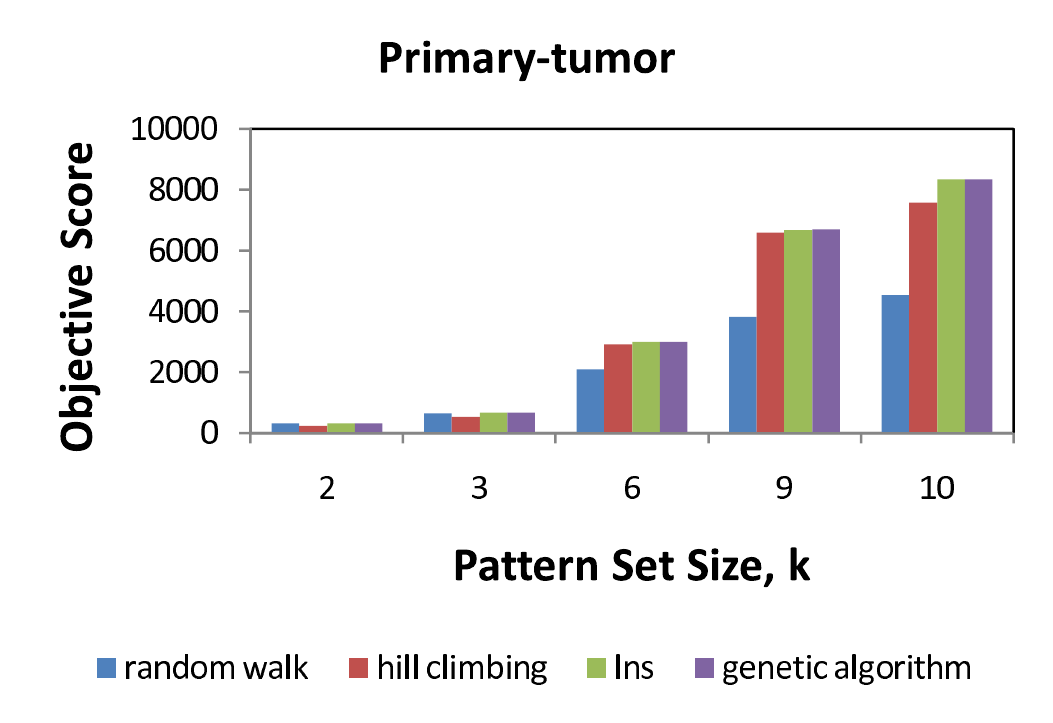}}
\caption{Bar diagram showing comparison of average objective score achieved by different
algorithms for various sizes of pattern sets, $k = 2, 3, 6, 9, 10.$}
\label{fig:2}
\end{figure}

\begin{figure}[h]
\centering
\subfigure[Average]{\includegraphics[width=0.49\textwidth]{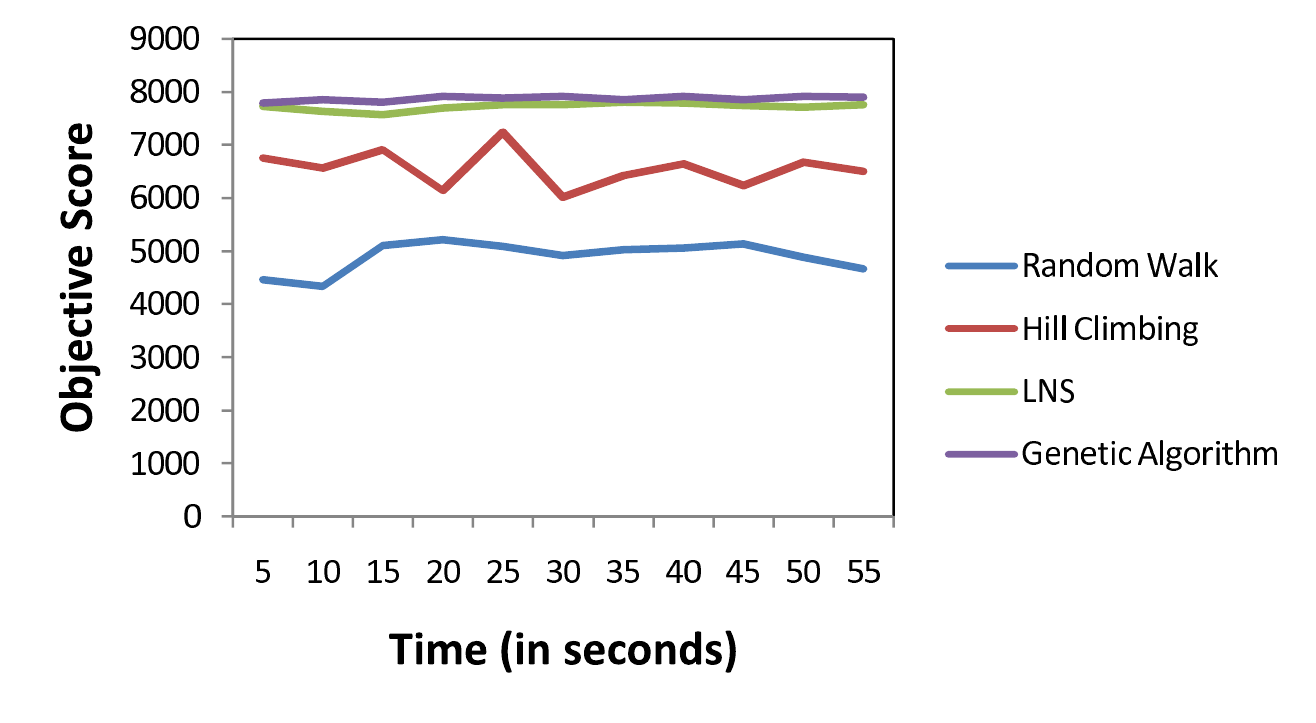}}
\subfigure[Best]{\includegraphics[width=0.49\textwidth]{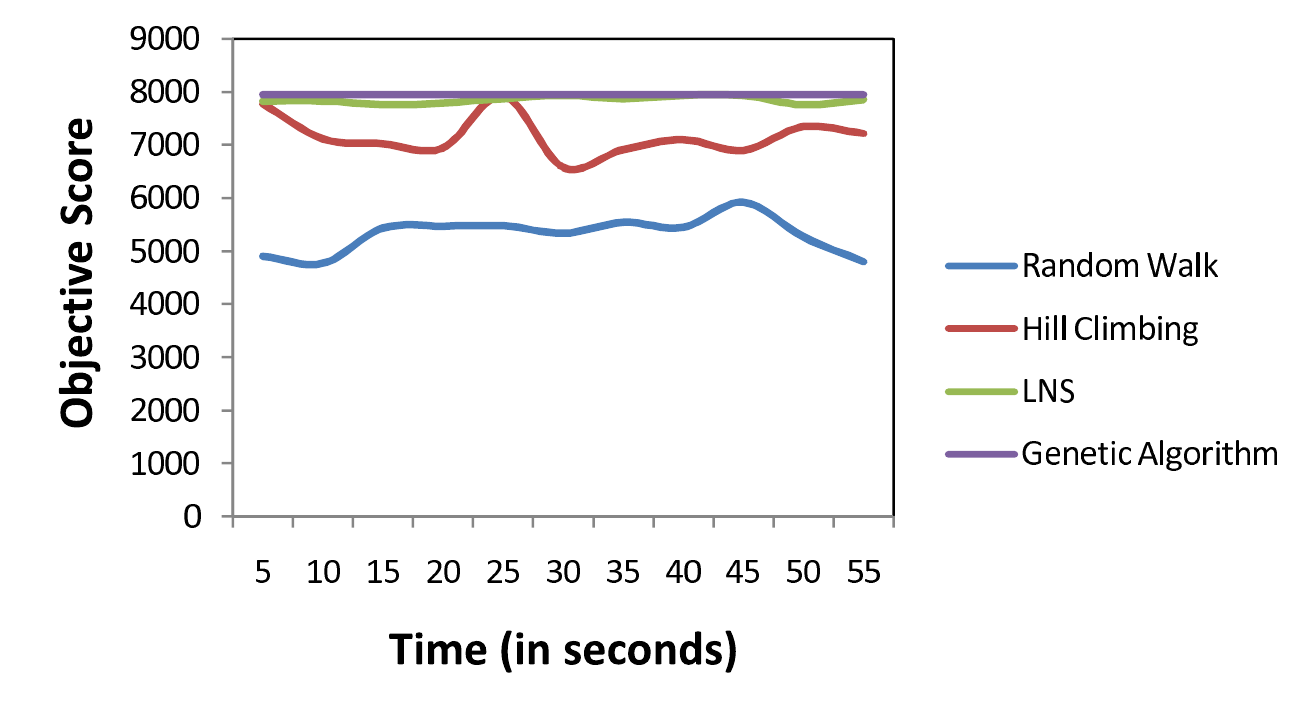}}
\caption{Comparison of objective score achieved by different
algorithms for the tic-tac-toe dataset with pattern size $k=6$. }
\label{fig:3}
\end{figure}

\section{Conclusion}

In this paper, we proposed a new genetic algorithm by tweaking (using random restart and twin removal along with mutation and crossover) to solve the task of mining diverse pattern sets. Here, genetic algorithm shows good results within a short period of time with compared to other algorithms. In future, we would like to improve the performance of the search techniques for genetic algorithm for large population size within the framework of stochastic local search and solve pattern set mining related problems with realistic datasets.
\bibliographystyle{IEEEtranS}
\bibliography{reference2}

\end{document}